\documentclass[conference]{IEEEtran}
\IEEEoverridecommandlockouts

\usepackage{cite}
\usepackage{amsmath,amssymb,amsfonts}
\usepackage{algorithmic}
\usepackage{graphicx}
\usepackage{textcomp}
\usepackage{xcolor}
\usepackage{enumitem}
\usepackage{tabularx}
\usepackage{multirow}
\usepackage{ulem} 
\def\BibTeX{{\rm B\kern-.05em{\sc i\kern-.025em b}\kern-.08em
    T\kern-.1667em\lower.7ex\hbox{E}\kern-.125emX}}
\begin{document}

\title{LLM-based Content Classification Approach for GitHub Repositories by the README Files
}

\author{\IEEEauthorblockN{1\textsuperscript{st} Malik Uzair Mehmood}
\IEEEauthorblockA{\textit{Department of Computing and Built Environment} \\
\textit{Birmingham City University}\\
Birmingham, England \\
malikuzair.mehmood@mail.bcu.ac.uk}
\and
\IEEEauthorblockN{2\textsuperscript{nd} Shahid Hussain}
\IEEEauthorblockA{\textit{Department of Computer Science and Software Engineering} \\
\textit{School of Engineering}\\
Penn State University, Behrend, Erie PA \\
shussain@psu.edu}
\and
\IEEEauthorblockN{3\textsuperscript{rd} Wen-Li Wang}
\IEEEauthorblockA{\textit{Department of Computer Science and Software Engineering} \\
\textit{School of Engineering}\\
Penn State University, Behrend, Erie PA \\
wxw18@psu.edu}
\and
\IEEEauthorblockN{\hspace{1.5cm}4\textsuperscript{th} Muhammad Usama Malik}
\IEEEauthorblockA{\textit{\hspace{1.5cm}Department of Electrical Engineering} \\
\textit{\hspace{1.5cm}UET Taxila}\\
\hspace{1.5cm}Rawalpindi, Pakistan \\
\hspace{1.5cm}usama.malik32@yahoo.com}

}

\maketitle

\begin{abstract}
GitHub is the world’s most popular platform for storing, sharing, and managing code. Every GitHub repository has a README file associated with it. The README files should contain project-related information as per the recommendations of GitHub to support the usage and improvement of repositories. However, GitHub repository owners sometimes neglected these recommendations. This prevents a GitHub repository from reaching its full potential. This research posits that the comprehensiveness of a GitHub repository's README file significantly influences its adoption and utilization, with a lack of detail potentially hindering its full potential for widespread engagement and impact within the research community. Large Language Models (LLMs) have shown great performance in many text-based tasks including text classification, text generation, text summarization and text translation. In this study, an approach is developed to fine-tune LLMs for automatically classifying different sections of GitHub README files. Three encoder-only LLMs are utilized, including BERT, DistilBERT and RoBERTa. These pre-trained models are then fine-tuned based on a gold-standard dataset consisting of 4226 README file sections. This approach outperforms current state-of-the-art methods and have achieved an overall F1 score of 0.98. Moreover, we have also investigated the use of Parameter-Efficient Fine-Tuning (PEFT) techniques like Low-Rank Adaptation (LoRA) and shown an economical alternative to full fine-tuning without compromising much performance. The results demonstrate the potential of using LLMs in designing an automatic classifier for categorizing the content of GitHub README files. Consequently, this study contributes to the development of automated tools for GitHub repositories to improve their identifications and potential usages.
\end{abstract}


\section{Introduction}
GitHub, with over 100 million contributive developers and more than 420 million repositories \cite{Kayle}, is the world’s largest platform for hosting open-source software projects, version control, and code sharing. Every GitHub repository holds a README file with it that describes several aspects of the project. GitHub in its official documentation \cite{githubReadme} provided its own set of guidelines for creating effective README files that include details like what is the function of the project, why the project is useful, how can someone use the project, where the help related to the project is available and who maintains the project. Tom Werner, the co-founder of GitHub coined the term Readme Driven Development (RDD) \cite{Werner2010} and emphasized that an excellent project with a poorly developed README file cannot be utilized to its full potential. 

\begin{figure}[b]
\centerline{\includegraphics[width=0.8\columnwidth]{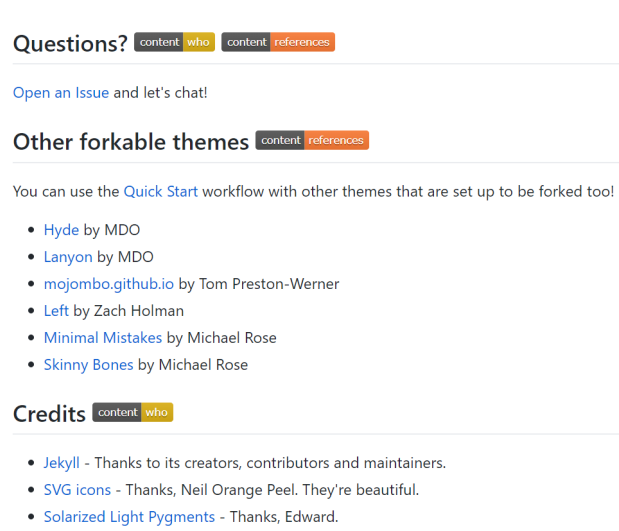}}
\caption{Labeled README File Sections}
\label{fig1}
\end{figure}

An open-source survey \cite{OpenSourceSurvey2017} conducted by GitHub shows that about 93\% of the respondents had shown their dissatisfaction with the quality of documentation and README files. There aren’t many studies that have addressed the content of GitHub README files to enhance the productivity of GitHub repositories. To address these challenges Prana et al \cite{prana2019categorizing}, conducted the first-ever study in which they created a gold standard dataset consisting of 4226 GitHub README file sections from 393 randomly sampled README files with the help of human labelers. A classifier was developed using 29 carefully curated heuristic features and achieved an F1 score of 0.746. This study also proved the practical effectiveness from 21 professionals who affirmed that labeling README file sections does improve the readability and quality of software repositories. Figure \ref{fig1} shows an example of a GitHub README with labeled file sections. However, the performance of this classifier is dependent on the availability of human labeled training data which is very expensive and time-consuming to produce.

In 2017, the introduction of Transformer \cite{vaswani2017attention} with ability to process sequential data through self-attention mechanisms overcame the diminishing gradient problem of recurrent neural networks. It revolutionized the current era of natural language processing, enabling unprecedented natural language understanding and generation capabilities by developing highly scalable large language models (LLMs), whose ability to perform text classification, summarization, generation, and translation advanced the way textual data was processed. Pre-trained LLMs on huge amounts of unlabeled data could be further fine-tuned and encoder-only transformer-based LLMs namely BERT \cite{devlin2018bert}, RoBERTa \cite{liu2019roberta} and DistilBERT \cite{sanh2019distilbert} have been proven to be effective for text classification tasks. This study aims to explore the potential of encoder-based LLMs to automate the task of classifying different sections of GitHub README files.  This article also reports the performance comparison of parameter-efficient fine-tuning (PEFT) vs full fine-tuning on GitHub README file sections classification. The followings lists the research questions (RQs) to be answered, 
\begin{enumerate}
\item RQ1: To what extent can the fine-tuned LLMs effectively categorize GitHub README file sections?
\item RQ2: Does PEFT produce comparable results to full fine-tuning for classifying GitHub README file sections?
\item RQ3: How well do LLMs methods compete with the existing state-of-the-art methods?
\end{enumerate}
The followings are the main contributions of this article, which
\begin{enumerate}
\item Achieved an F1 score of 0.98 surpassing the state-of-the-art results by a significant margin.
\item Compared the performance of supervised learning methods with that of LLMs for classifying file sections of GitHub README files.
\item Demonstrated the effectiveness of PEFT techniques of LLMs vs full fine-tuning.
\end{enumerate}

The rest of the article is organized as follows, Section II presents the recent literature related to GitHub documentation and text classification, Section III discusses the adopted research methodologies for different experiments, Section IV presents the results of those experiments, Section V describes the threats to validity for this study, Section VI shows the implications of this study for the research community, and finally Section VII concludes the article with highlighting future directions.
\section{Related Work}
Text classification is a task of natural language processing in which a given text is classified into one of the several classes. In our case, we have performed multi-label text classification in which each GitHub README file section is classified into one or more classes at the same time. Therefore, this section starts with a quick overview of text classification followed by a quick overview of work being done on GitHub README files.

\subsection{Brief history of text classification and natural language processing}
Text classification has been studied and addressed in many real-world applications in several domains including health \cite{zhang2018patient2vec}, law \cite{turtle1995text}, social sciences \cite{ofoghi2017textual}, software engineering \cite{watanabe2020reducing}, and businesses \cite{yu2011classifying} over the past few decades. The study \cite{kowsari2019text} shows that in a nutshell, text classification tasks revolve around four key phases namely feature extraction, dimensionality reduction, selection of suitable classifier, and evaluation strategies. The types of algorithms and approaches used for text classification have evolved over time, starting with a rule-based approach \cite{frase1998computer} to process textual data in which computer programs were written based on predefined rules to perform text processing. Moving on to Hidden Markov Models (HMMs) \cite{eddy2004hidden}\cite{kang2018opinion}, the approach was seen as a new solution to the text classification problem through statistical techniques. For small size datasets HMMs gained promising results. With the growing popularity of machine learning (ML), supervised algorithms including Support Vector Machines \cite{tong2001support}, K-Nearest Neighbor \cite{yong2009improved}, Random Forest \cite{chaudhary2016improved}, Logistic Regression \cite{ifrim2008fast}, and more, were then used to support text classification tasks. The application of these methods however depended on the availability of labeled datasets, i.e., expensive to get. As the size and availability of datasets continue to grow, deep learning took over. Many deep learning algorithms including Recurrent Neural Networks (RNNs) \cite{lai2015recurrent}, Convolutional Recurrent Neural Networks (CRNN) \cite{wang2019convolutional}, Long Short Term Memory (LSTM) \cite{lyu2021convolutional} and Gated Recurrent Unit (GRU) \cite{zulqarnain2020text} became dominant in the field.

In 2017, the introduction of Transformers \cite{vaswani2017attention} changed the landscape of text processing. The transformer architecture of Large Language Models (LLMs) took the world by storm. it came up with a very simple encoder-decoder structure and self-attention mechanism that revolutionized the NLP industry. Unlike RNN, the LLMs were comparatively easier to parallelize, more scalable, and flexible to be used on several different tasks \cite{zhao2023survey}. The power of Transformer architectures can leverage self-attention mechanisms and parallelization capabilities to process and generate vast amounts of sequential data, thereby achieving state-of-the-art results in various natural language processing tasks. Encoder-only architecture has proven to be effective for tasks like text classification, sentiment analysis, and Entity Name Recognition. In contrast, decoder-only architecture is better suited for text summarization, text generation, and question-answering. Additionally, encoder-decoder architecture is more suitable to tasks like text translation, language translation, etc. LLMs offer the ability to use models pre-trained on massively huge amounts of data for simpler tasks through fine-tuning a very small amount of labeled data. This opens so many possibilities and use cases that could be exploited. Fine-tuning all parameters is a resource-intensive task. In this study Parameter Efficient Fine Tuning (PEFT) technique \cite{han2024parameter} Low-Rank Adaptation (LoRA) is applied to simplify the task of fine-tuning under fewer resources.

Besides, the Encoder-only models including BERT, RoBERTa and DistilBERT are used to perform text classification on GitHub README file sections to improve the quality, readability, and potential of GitHub repositories.
\subsection{Historical work done on GitHub README files}
In 2019 Prana et. al. \cite{prana2019categorizing}. conducted a qualitative study on understanding the content of a typical GitHub README file. Their study contributed a gold standard baseline dataset annotated by humans consisting of 8 different labels namely what, why, how, when, who, references, contributions, and others class.  The study also designed a classifier to automatically categorize the content of GitHub README files using statistical and heuristical features and achieved a weighted F1 score of 0.746. Subject to limited amount of data, results could not be further improved as it is expensive and time-consuming to produce a bigger labeled dataset. In \cite{wang2023study}, Wang et.al. studied the relationship between the popularity of a GitHub repository and associated README file. They concluded that a high-quality GitHub README file contributes heavily to the popularity of the repository. The authors recommended that repository owners should invest time in documenting their project with proper formatting, detailed instructions and continuous updates to keep the project alive. In another study, Gao et al. \cite{gao2023evaluating} evaluated the use of transfer learning for simplifying GitHub README files using transformer-based approach. The authors collected a dataset consisting of 14588 pairs of README files and trained a transformer-based model to effectively summarize complex ones that contained technical jargon. This study also highlighted the importance of a good README file for a successful repository. In a related study \cite{doan2023too}, a transformer-based approach was conducted to summarize GitHub README files. Authors used two transformer-based models namely BART and T5 to summarize the contents and evaluated their results using ROGUE-1, ROGUE-2 and ROGUE-L metrics. Their results again highlighted the importance of a thorough README file. In a similar study, Liu et. al. \cite{liu2022readme} applied statistical analyses and clustering methods on 14901 Java repositories and highlighted the association between file structures and the popularity of the repositories.  

With the confidence on the importance of the  GitHub README files, the aim of this study is to classify the file contents to help developers better understand the purpose of these files without spending too much time looking for the needed information.
\section{METHODOLOGY}
\begin{figure}[t]
\centerline{\includegraphics[width=0.9\columnwidth]{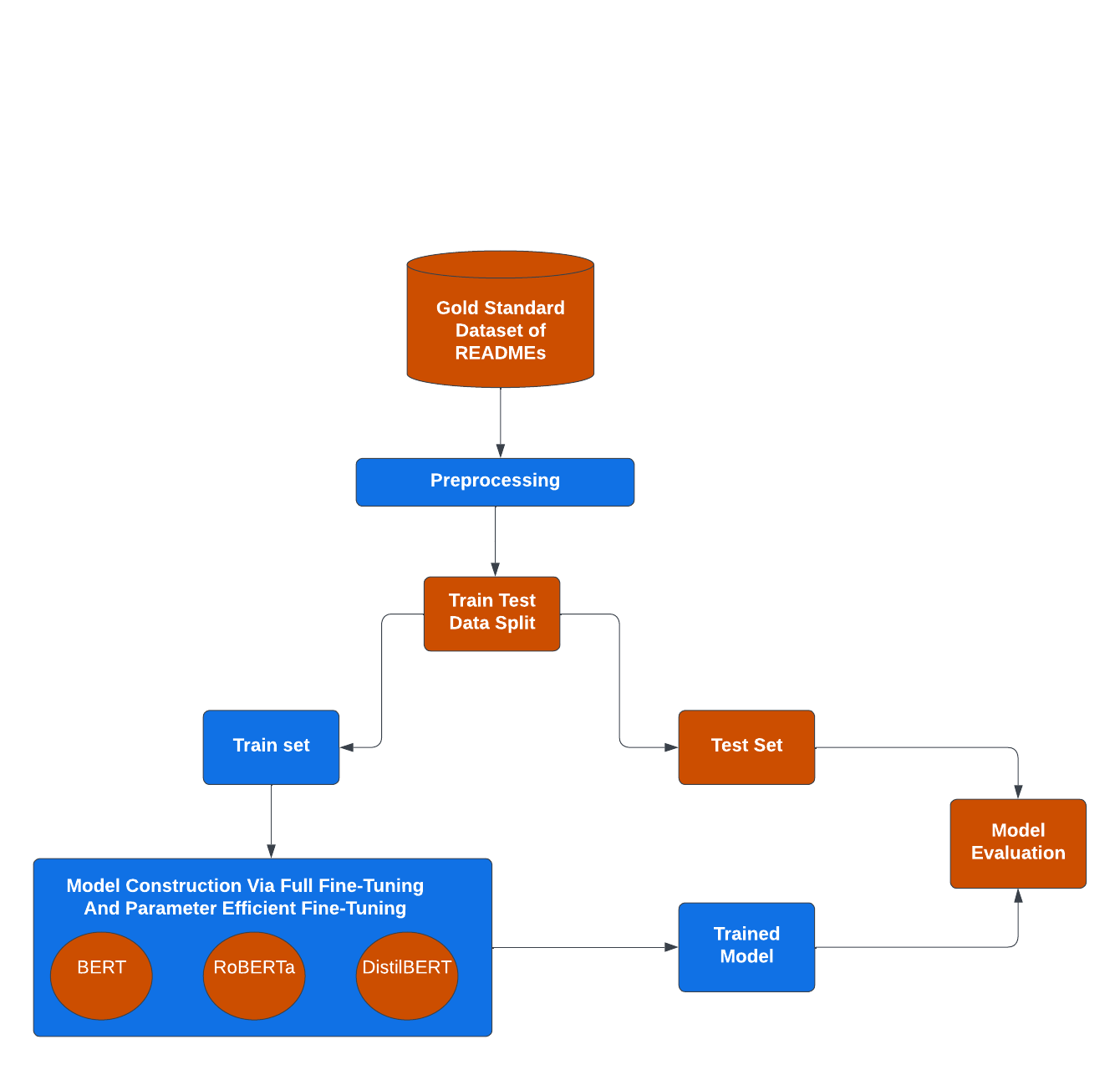}}
\caption{Proposed Methodology}
\label{fig2}
\end{figure}

This section illustrates each step in our proposed methodology, which is depicted by a high-level diagram as shown in
Figure \ref{fig2}.
\subsection{Dataset}
\begin{figure}[t]
\centerline{\includegraphics[width=0.9\columnwidth]{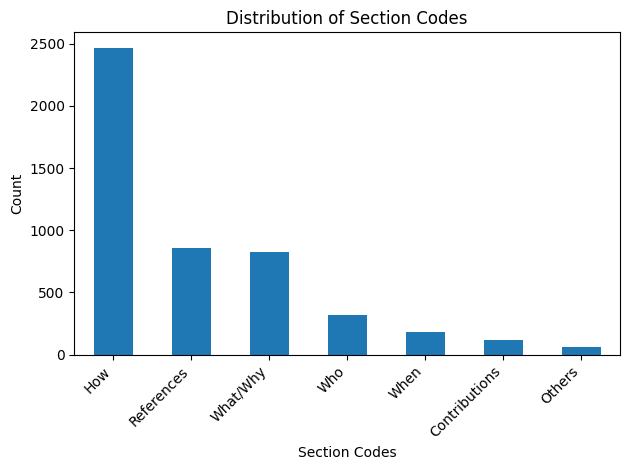}}
\caption{File Section Class wise Distribution}
\label{fig3}
\end{figure}

This study incorporates a gold standard humanly annotated dataset \cite{prana2019categorizing} that was collected randomly from GitHub. The dataset consists of 393 README files and 4226 file sections. There are a total of 8 classes, and the details are given in Table \ref{my_table}. Each file section could belong to more than one class at a time. This is an imbalanced dataset with different proportions of labels for different classes. Figure \ref{fig3} shows a distribution of each class present in the dataset. Due to an insignificant amount in the what and why classes, they are merged into one class. This dataset involves README files that are in English language only with file size $>$ 2KB. GitHub README files are written in a special format called markdown which follows an HTML-like structure i.e. h1 to h6 for different heading levels which are exploited to extract sections from a file. Each section consists of a heading and associated text. 
\begin{table}[b]
\centering
\caption{Class Label Description \cite{prana2019categorizing}}
\label{my_table}
\begin{tabular}{cp{6cm}}
\hline
Class Label & Example Section Heading \\
\hline
What & Introduction, project background \\
Why & Advantages of the project, comparison with related work \\
How & Getting started/quick start, how to run, installation, how to update, configuration, setup, requirements, dependencies, languages, platforms, demo, downloads, errors and bugs \\
When & Project status, versions, project plans, roadmap \\
Who & Project team, community, mailing list, contact, acknowledgment, license, code of conduct \\
References & API documentation, getting support, feedback, more information, translations, related projects \\
Contributions &  \\
Other & Contributing guidelines \\
\hline
\end{tabular}
\end{table}
\subsection{Data Pre-processing}
\begin{figure}[t]
\centerline{\includegraphics[width=0.9\columnwidth]{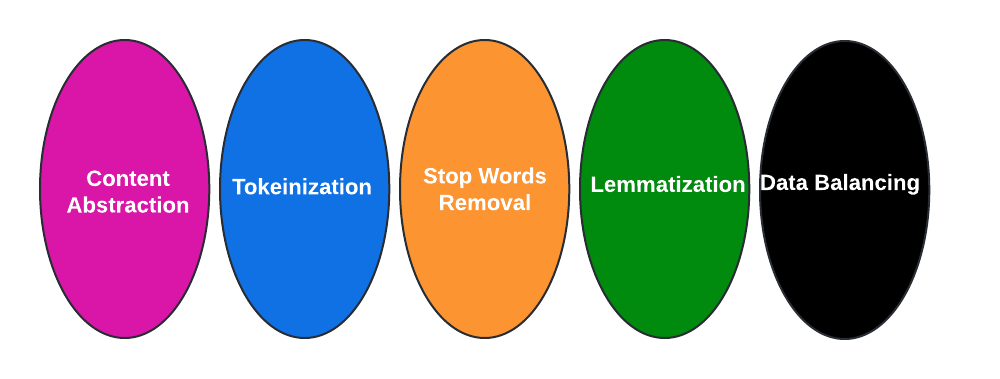}}
\caption{Data Preprocessing}
\label{fig4}
\end{figure}

A series of prep-processing steps are performed on each file section which are depicted in Figure \ref{fig4} and explained below.
\subsubsection{Content Abstraction}
Eight different types of content from each README file section are extracted namely code blocks, tables, mailto links, images, unordered lists, ordered lists, hyperlinks, and numbers. Each one of these is replaced with placeholder values that are mentioned in Table \ref{my_placeholders}. The idea behind this approach is that the presence of a code block or table in a section is enough for the classifier training regardless of what the value of the content is. 
\begin{table}[b]
\centering
\caption{Content Placeholders}
\label{my_placeholders}
\begin{tabular}{cp{6cm}}
\hline
Content Type & Content Placeholders \\
\hline
Code Tags & CODE \\
Hyper Links & ANCHOR \\
Tables & TABLE \\
Image Tags & IMAGE \\
Ordered Lists & OL \\
Unordered Lists & UL \\
Mail to Links & MAILTO \\
Numbers & NUMBER \\
\hline
\end{tabular}
\end{table}
\subsubsection{Tokenization}
In this step, data is tokenized to generate word embeddings to be used in our model. This step is crucial because models cannot process plain text; therefore, the data is first converted into tokens, and then word embeddings are created from these tokens. Various tokenizers will be employed in our experiments, depending on the selected pre-trained model.
\subsubsection{Stop Words Removal}
Stop words are removed, e.g., the, and, of, to etc. from file sections that contribute very little to the semantic meaning of the sentence. This will have the pre-trained models focus on more informative words and achieve better performance. The NLTK library \cite{hardeniya2016natural} helps achieve this task.
\subsubsection{Lemmatization}
It is performed on the data to simplify words into its base form. This plays an important role in normalization of data and enhances semantic analysis capability of the models. We used WordNetLemmatizer from NLTK facilitates this task.
\subsubsection{Data Balancing}
An imbalanced dataset often results in overfitting. In our case, the dataset is imbalanced with more instances in the \textit{how} class but few instances in the \textit{other} and \textit{contribution} classes. Hence, the dataset is improved by oversampling to match the minority classes with the number of instances in the majority class. 
\subsection{Data Splitting}
This step applies a 70/30 train/test split on the dataset to train and test our model. Stratified sampling to make sure that class distribution is balanced in both the test and train sets. Additionally, k-fold cross-validation is applied to validate the results.
\subsection{Model Construction}
A fine-tuning is done to three pre-trained models using two different approaches. The first approach starts with full fine-tuning and the other is by adapter-based parameter efficient fine-tuning using Low Rank Adaptation (LoRA). Details of both approaches are given below.
\subsubsection{Full Fine-tuning}
In the experiments, different pre-trained encoder-only language models are used to perform fine-tuning. They are described as below.
\paragraph{BERT (Bidirectional Encoder Representations from Transformers)}
The bert-base-uncased pre-trained model is used to fine-tune our perform fine-tuning our README dataset. BERT base is pre-trained on 110 million parameters and has a maximum capacity of 512 tokens to be processed at the same time. BERT uses Masked Language Modeling (MLM) which masks 15\% of the input tokens and trains the model to predict the masked words based on the context. BERT consists of several encoding layers that use a self-attention mechanism and a feed-forward neural network. The output of the final encoder layer is used for the downstream of our README file section classification task. Because the tokenized size of our README file section can vary padding and truncation are used in our model to adjust and make consistent the tokenzied size according to the model limitations. Moreover, an early stopping and dropout layer is applied during fine-tuning to deal with the overfitting problem and different parameters are experimented including including different numbers of epochs, learning rates, and devices per batch.
\paragraph{RoBERTa (Robustly Optimized BERT)}
RoBERTa is an optimized form of BERT and it is also pre-trained on 110 million parameters. The RoBERTa is another base model used to fine-tune our task. Just like BERT it has a  transformer-based architecture, but uses a dynamic masking approach in which tokens are masked randomly on each epoch instead of static masking.
\paragraph{DistilBERT (Distilled BERT)}
DistilBERT is a distilled version of BERT and is created using knowledge distillation techniques. This version of BERT is typically pretrained on 66 million parameters which are significantaly less than the BERT model. The model offers an advantage of performing fine-tuning on our dataset for it being comparatively more economical in terms of size and memory requirements, hence faster as well. DistilBERT proves its effectiveness in showing results for multi label classification problems based on the input of our README file sections.
\subsubsection{Parameter Efficient Fine-Tuning (PEFT)}
To reduce the burden of full fine-tuning, the adapter based PEFT technique is applied to those three pre-trained models for our task to seek improvements. Instead of updating all parameters on the downstrean task for full fine-tuning, the PEFT technique will only update a subset of parameters to save both time and memory and also ensure the model retains its pre-trained form. This can prevent catastrophic forgetting [32] in which a model forgets its pretraining and is only good for the targeted class. LoRA (Low-Rank Adaptation) is utilized in this case. Details of which are given below.
\paragraph{LoRA (Low Ranked Adaptation)}
LoRA \cite{hu2021lora} is an adapter-based PEFT technique. In this technique rather than updating all parameters of the pre-trained models, we add adapters to the pre-trained layers which are then fine-tuned to meet our required task. This can be achieved by adding rank-decomposition matrices to different layers of the model. The approach is computationally less intensive as compared to the full fine-tuning. Equation \ref{LoRA Equation} formulates the computation of LoRA. 
\begin{equation} 
W' = W + AB^T\label{LoRA Equation}
\end{equation}
where \textit{W} is the original weight matrix of the layer with a full dimension of ($\alpha$,$\alpha$), \textit{A} is the low-rank matrix with a dimension of ($\alpha$,r), $B^\top T$ is another low-rank matrix transposed to a dimension of (r,$\alpha$), and r is the rank of decomposition matrix where r \textless $\alpha$.  By training only the \textit{A} and \textit{B} matrices, we can effectively adapt the model to the downstream task by modifying only a subset of original weights of the pre-trained model. 
Table \ref{FullvsLoRA} shows a comparison of a number of the number of parameters that are updated in full fine-tuning vs PEFT. 
\begin{table}[t]
\centering
\caption{Full Fine-tuning vs LoRA Fine-tuning (Updated Parameters)}
\label{FullvsLoRA}
\begin{tabular}{|c|c|c|}
\hline
Model Name & Full Fine-tuning & LoRA Fine-tuning (r=4) \\
\hline
BERT Base & 109,566,734 & 79,111 \\
\hline
RoBERTa Base & 125,320,718 & 669,703 \\
\hline
DistilBERT & 67,591,694 & 632,839 \\
\hline
\end{tabular}
\end{table}
\subsection{Evaluation Criteria}
Several standard matrices are adopted in our multi-label classification problems including F1 score, ROC AUC, Mathew Correlation Coefficient (MCC), and Kappa score to evaluate the results. 
\subsection{Experimental Setup}
A variety of experiments were conducted based on the Google Colab T4 GPU to speed up the fine-tuning of our adopted models. During full fine-tuning, several hyperparameters were exercised including learning rate, number of epochs, dropout layers to overcome overfitting, and different numbers of batch sizes while training. The experimental setups were different for different models and the best achieved results under different setups were observed and recorded. For the train/test sets a 70/30 split with stratified sampling to validate the results.

Likewise, during PEFT techniques, several hyperparameters were also experimented. By Equation 1, the rank r that determines the dimensionality of low rank matrices A and B, and the alpha a that scales the contribution of low-rank matrices back to the original weights were two main hyper-parameters to be exervised while working with LoRA. Diverse settings were experimented with the best results to be reported. 
\begin{table*}[t]
\centering
\caption{Transformer based models comparison with state-of-the-art techniques}
\label{TransformervsSOTA}
\begin{tabular}{c|c|c|c|c}
\hline
Model/Technique & F1 Score & ROC AUC & MCC & Kappa Score \\
\hline
SVM\cite{prana2019categorizing} & 0.746 & 0.957 & 0.844 & 0.831 \\
Random Forest & 0.696 & 0.929 & 0.848 & 0.841 \\
Naive Bayes & 0.518 & 0.832 & 0.662 & 0.747 \\
BERT Base & \textbf{0.980} & \textbf{0.989} & \textbf{0.974} & \textbf{0.975} \\
RoBERTa Base& 0.966 & 0.982 & 0.955 & 0.956 \\
DistilBERT & 0.958 & 0.963 & 0.955 & 954 \\

\hline
\end{tabular}
\end{table*}
\begin{table*}[t]
\centering
\caption{Full Fine Tuning vs LoRA Fine Tuning}
\label{FullvsLoRAFT}
\begin{tabular}{c|cccc|cccc}
\hline
\multirow{2}{*}{Model/Technique} & \multicolumn{4}{c|}{Full Fine Tuning} & \multicolumn{4}{c}{LoRA Fine Tuning} \\
\cline{2-9}
 & F1 Score & ROC AUC & MCC & Kappa Score & F1 Score & ROC AUC & MCC & Kappa Score \\
\hline
BERT Base & 0.980 & 0.989 & 0.974 & 0.975 & 0.860 & 0.928 & 0.804 & 0.797 \\
RoBERTa Base & 0.966 & 0.982 & 0.955 & 0.956 & 0.892 & 0.946 & 0.851 & 0.851 \\
DistilBERT & 0.958 & 0.963 & 0.955 & 0.954 & 0.908 & 0.953 & 0.871 & 0.870 \\

\hline
\end{tabular}
\end{table*}
\subsection{RESULTS \& DISCUSSIONS}
Tables \ref{TransformervsSOTA} and \ref{FullvsLoRAFT} show the results in two different perspectives. Table \ref{TransformervsSOTA} shows a comparison of our fine-tuned models including BERT, RoBERTa and DistilBERT with the state-of-the-art results reported in literature using SVM, Random Forest, and Naive Bayes classifiers, whereas \ref{FullvsLoRAFT} shows the comparison of full fine tuning against PEFT.

To answer RQ1 on the effectiveness of LLMs in performing GitHub README file sections classification we fine-tuned BERT, RoBERTa and DistilBERT encoder only models. For computing MCC, Kappa Score, F1-score and ROC AUC we used labels predicted by our fine tuned models. For ground truth labels we used the original labels from our test dataset. Our results show that all evaluation criteria were scored over 0.9 that shows our models have shown strong performance in accomplishing the desired task. 

To answer RQ2 on the performance of PEFT vs full fine tuning evaluate the performance of parameter efficient fine tuning compared to full fine tuning, Table \ref{FullvsLoRAFT} shows that cost saving PEFT techniques were still able to produce pretty close results to full fine-tuning. In particular, the LoRA based DistilBERT PEFT technique yielded a superior result than other tested models models with a F1 score of 0.908. Although the performance of full fine-tuning is hard to beat due to its advantage of learning intricate details from larger models, the significantly less number of parameters to be updated by PEFT proves its huge saving of computational resources to train the model. To answer RQ3, the results in table IV show that the LLM-based fine-tuning models outperformed the traditional machine learning algorithms in all evaluation metrics including F1 score, ROC AUC, MCC, and Kappa Score. This concludes that those pre-trained models can classify GitHub README file sections more effectively. 

\section{THREATS TO VALIDITY}
There are a few limitations to this study. Firstly, the dataset used in this study is imbalanced and needs data resampling techniques to be applied that may cause bias wjile training. Secondly, the reported study is based on English text only and does not account for a multilingual dataset. Future studies with larger labeled dataset and multilingual pre-trained models should be warranted to address these issues.  
\section{IMPLICATIONS TO RESEARCH COMMUNITY}
The results of this study contribute to the growing body of research on LLMs and  the transformer based pre-trained models. By utilizing encoder only transformer models for classifying GitHub README file sections, this research provided valuable insights into the potential of LLMs for text classification tasks. Moreover, this study also provides insights into the effectiveness of using PEFT methods in comparison to full fine tuning techniques. Future research can explore the utilization of LLMs on multilingual GitHub README files. The proposed methodology can also be extended to other code sharing and version control platforms.   
\subsection{CONCLUSION}
In this study we have investigated the potential of using encoder only LLMs including BERT, RoBERTA and DistilBERT for classifying GitHub README file sections into several classes. The results show that these transformer based methods clearly outperform the current state of the art (SOTA) models. Two types of experiments were set up and performed. The first type is to fully fine-tune the pre-trained models, while the second one adopts parameter efficient fine tuning (PEFT) to conserve time and cost. In both cases, all these fine-tuned models were able to produce better results as compared to the SOTA models. The comparisons between the full and PEFT models show that full fine tuning is more accurate while PEFT is much less computation intensive. Most importantly, the PEFT fine-tuned models still yield better results than those of SOTA models. This study concludes that LLMs have a huge potential in improving the software engineering domain and can be adapted to support more platforms after fine-tuning. 
For the future work, one direction is to increase the size of the gold standard dataset and develop a pipeline that is applicable to other code sharing platforms. Another direction is to incorporate these successful models into tools to help practitioners automatically cateforize the content of GitHub README files In future one approach would be to increase the size of gold standard dataset and develop a pipeline that could be applied to any code sharing platform. Moreover, these models could also be utilized to develop tools that could automatically categories the content of GitHub README files. Full code of our experiments is publicaly available at https://github.com/uzair-malik/LLM-based-GitHub-README-File-Content-Classification.

\bibliography{octaverse}

\begin{thebibliography}{10}
\providecommand{\url}[1]{#1}
\csname url@samestyle\endcsname
\providecommand{\newblock}{\relax}
\providecommand{\bibinfo}[2]{#2}
\providecommand{\BIBentrySTDinterwordspacing}{\spaceskip=0pt\relax}
\providecommand{\BIBentryALTinterwordstretchfactor}{4}
\providecommand{\BIBentryALTinterwordspacing}{\spaceskip=\fontdimen2\font plus
\BIBentryALTinterwordstretchfactor\fontdimen3\font minus \fontdimen4\font\relax}
\providecommand{\BIBforeignlanguage}[2]{{%
\expandafter\ifx\csname l@#1\endcsname\relax
\typeout{** WARNING: IEEEtran.bst: No hyphenation pattern has been}%
\typeout{** loaded for the language `#1'. Using the pattern for}%
\typeout{** the default language instead.}%
\else
\language=\csname l@#1\endcsname
\fi
#2}}
\providecommand{\BIBdecl}{\relax}
\BIBdecl

\bibitem{Kayle}
\BIBentryALTinterwordspacing
K.~Daigle and G.~Staff, ``Octoverse: The state of open source and rise of ai in 2023.'' [Online]. Available: \url{https://github.blog/tag/research/}
\BIBentrySTDinterwordspacing

\bibitem{githubReadme}
\BIBentryALTinterwordspacing
GitHub, ``About readmes.'' [Online]. Available: \url{https://docs.github.com/en/repositories/managing-your-repositorys-settings-and-features/customizing-your-repository/about-readmes}
\BIBentrySTDinterwordspacing

\bibitem{Werner2010}
\BIBentryALTinterwordspacing
T.~Werner, ``Readme driven development,'' Blog post, 2010. [Online]. Available: \url{https://tom.preston-werner.com/2010/08/23/readme-driven-development.html}
\BIBentrySTDinterwordspacing

\bibitem{OpenSourceSurvey2017}
\BIBentryALTinterwordspacing
O.~S.~S. Team, ``{Open Source Survey}.'' [Online]. Available: \url{https://opensourcesurvey.org/2017/}
\BIBentrySTDinterwordspacing

\bibitem{prana2019categorizing}
G.~A.~A. Prana, C.~Treude, F.~Thung, T.~Atapattu, and D.~Lo, ``Categorizing the content of github readme files,'' \emph{Empirical Software Engineering}, vol.~24, pp. 1296--1327, 2019.

\bibitem{vaswani2017attention}
A.~Vaswani, ``Attention is all you need,'' \emph{arXiv preprint arXiv:1706.03762}, 2017.

\bibitem{devlin2018bert}
J.~Devlin, M.-W. Chang, K.~Lee, and K.~Toutanova, ``Bert: Bidirectional encoder representations from transformers,'' \emph{arXiv preprint arXiv:1810.04805}, 2018.

\bibitem{liu2019roberta}
Y.~Liu, M.~Ott, N.~Goyal, J.~Du, M.~Joshi, D.~Chen, O.~Levy, M.~Lewis, L.~Zettlemoyer, and V.~Stoyanov, ``Roberta: A robustly optimized bert pretraining approach,'' \emph{arXiv preprint arXiv:1907.11692}, 2019.

\bibitem{sanh2019distilbert}
V.~Sanh, L.~Debut, J.~Chaumond, and T.~Wolf, ``Distilbert, a distilled version of bert: smaller, faster, cheaper and lighter,'' \emph{arXiv preprint arXiv:1910.01108}, 2019.

\bibitem{zhang2018patient2vec}
J.~Zhang, K.~Kowsari, J.~H. Harrison, J.~M. Lobo, and L.~E. Barnes, ``Patient2vec: A personalized interpretable deep representation of the longitudinal electronic health record,'' \emph{IEEE Access}, vol.~6, pp. 65\,333--65\,346, 2018.

\bibitem{turtle1995text}
H.~Turtle, ``Text retrieval in the legal world,'' \emph{Artificial Intelligence and Law}, vol.~3, pp. 5--54, 1995.

\bibitem{ofoghi2017textual}
B.~Ofoghi and K.~Verspoor, ``Textual emotion classification: An interoperability study on cross-genre data sets,'' in \emph{Australasian Joint Conference on Artificial Intelligence}.\hskip 1em plus 0.5em minus 0.4em\relax Springer, 2017, pp. 262--273.

\bibitem{watanabe2020reducing}
W.~M. Watanabe, K.~R. Felizardo, A.~Candido~Jr, {\'E}.~F. de~Souza, J.~E. de~Campos~Neto, and N.~L. Vijaykumar, ``Reducing efforts of software engineering systematic literature reviews updates using text classification,'' \emph{Information and Software Technology}, vol. 128, p. 106395, 2020.

\bibitem{yu2011classifying}
B.~Yu and L.~Kwok, ``Classifying business marketing messages on facebook,'' \emph{Proceedings of the Association for Computing Machinery Special Interest Group on Information Retrieval, Bejing, China}, pp. 24--28, 2011.

\bibitem{kowsari2019text}
K.~Kowsari, K.~Jafari~Meimandi, M.~Heidarysafa, S.~Mendu, L.~Barnes, and D.~Brown, ``Text classification algorithms: A survey,'' \emph{Information}, vol.~10, no.~4, p. 150, 2019.

\bibitem{frase1998computer}
L.~T. Frase, J.~Faletti, A.~Ginther, and L.~Grant, ``Computer analysis of the toefl test of written english,'' \emph{ETS Research Report Series}, vol. 1998, no.~2, pp. i--26, 1998.

\bibitem{eddy2004hidden}
S.~R. Eddy, ``What is a hidden markov model?'' \emph{Nature biotechnology}, vol.~22, no.~10, pp. 1315--1316, 2004.

\bibitem{kang2018opinion}
M.~Kang, J.~Ahn, and K.~Lee, ``Opinion mining using ensemble text hidden markov models for text classification,'' \emph{Expert Systems with Applications}, vol.~94, pp. 218--227, 2018.

\bibitem{tong2001support}
S.~Tong and D.~Koller, ``Support vector machine active learning with applications to text classification,'' \emph{Journal of machine learning research}, vol.~2, no. Nov, pp. 45--66, 2001.

\bibitem{yong2009improved}
Z.~Yong, L.~Youwen, and X.~Shixiong, ``An improved knn text classification algorithm based on clustering,'' \emph{Journal of computers}, vol.~4, no.~3, pp. 230--237, 2009.

\bibitem{chaudhary2016improved}
A.~Chaudhary, S.~Kolhe, and R.~Kamal, ``An improved random forest classifier for multi-class classification,'' \emph{Information Processing in Agriculture}, vol.~3, no.~4, pp. 215--222, 2016.

\bibitem{ifrim2008fast}
G.~Ifrim, G.~Bakir, and G.~Weikum, ``Fast logistic regression for text categorization with variable-length n-grams,'' in \emph{Proceedings of the 14th ACM SIGKDD international conference on Knowledge discovery and data mining}, 2008, pp. 354--362.

\bibitem{lai2015recurrent}
S.~Lai, L.~Xu, K.~Liu, and J.~Zhao, ``Recurrent convolutional neural networks for text classification,'' in \emph{Proceedings of the AAAI conference on artificial intelligence}, vol.~29, no.~1, 2015.

\bibitem{wang2019convolutional}
R.~Wang, Z.~Li, J.~Cao, T.~Chen, and L.~Wang, ``Convolutional recurrent neural networks for text classification,'' in \emph{2019 international joint conference on neural networks (IJCNN)}.\hskip 1em plus 0.5em minus 0.4em\relax IEEE, 2019, pp. 1--6.

\bibitem{lyu2021convolutional}
S.~Lyu and J.~Liu, ``Convolutional recurrent neural networks for text classification,'' \emph{Journal of Database Management (JDM)}, vol.~32, no.~4, pp. 65--82, 2021.

\bibitem{zulqarnain2020text}
M.~Zulqarnain, R.~Ghazali, Y.~M. Hassim, and M.~Rehan, ``Text classification based on gated recurrent unit combines with support vector machine,'' \emph{International Journal of Electrical and Computer Engineering}, vol.~10, no.~4, pp. 3734--3742, 2020.

\bibitem{zhao2023survey}
W.~X. Zhao, K.~Zhou, J.~Li, T.~Tang, X.~Wang, Y.~Hou, Y.~Min, B.~Zhang, J.~Zhang, Z.~Dong \emph{et~al.}, ``A survey of large language models,'' \emph{arXiv preprint arXiv:2303.18223}, 2023.

\bibitem{han2024parameter}
Z.~Han, C.~Gao, J.~Liu, S.~Q. Zhang \emph{et~al.}, ``Parameter-efficient fine-tuning for large models: A comprehensive survey,'' \emph{arXiv preprint arXiv:2403.14608}, 2024.

\bibitem{wang2023study}
T.~Wang, S.~Wang, and T.-H.~P. Chen, ``Study the correlation between the readme file of github projects and their popularity,'' \emph{Journal of Systems and Software}, vol. 205, p. 111806, 2023.

\bibitem{gao2023evaluating}
H.~Gao, C.~Treude, and M.~Zahedi, ``Evaluating transfer learning for simplifying github readmes,'' in \emph{Proceedings of the 31st ACM Joint European Software Engineering Conference and Symposium on the Foundations of Software Engineering}, 2023, pp. 1548--1560.

\bibitem{doan2023too}
T.~T. Doan, P.~T. Nguyen, J.~Di~Rocco, and D.~Di~Ruscio, ``Too long; didn’t read: Automatic summarization of github readme. md with transformers,'' in \emph{Proceedings of the 27th International Conference on Evaluation and Assessment in Software Engineering}, 2023, pp. 267--272.

\bibitem{liu2022readme}
Y.~Liu, E.~Noei, and K.~Lyons, ``How readme files are structured in open source java projects,'' \emph{Information and Software Technology}, vol. 148, p. 106924, 2022.

\bibitem{hardeniya2016natural}
N.~Hardeniya, J.~Perkins, D.~Chopra, N.~Joshi, and I.~Mathur, \emph{Natural language processing: python and NLTK}.\hskip 1em plus 0.5em minus 0.4em\relax Packt Publishing Ltd, 2016.

\bibitem{hu2021lora}
E.~J. Hu, Y.~Shen, P.~Wallis, Z.~Allen-Zhu, Y.~Li, S.~Wang, L.~Wang, and W.~Chen, ``Lora: Low-rank adaptation of large language models,'' \emph{arXiv preprint arXiv:2106.09685}, 2021.

\end{thebibliography}

\vspace{12pt}

\end{document}